# HUMAN ACTIVITY RECOGNITION USING DEEP LEARNING APPROACHES: SINGLE FRAME CNN AND CONVOLUTIONAL LSTM

**Sheryl Mathew, Annapoorani Subramanian, S. Pooja, Balamurugan MS**

**School of Electronics Engineering, Vellore Institute of Technology, Chennai, India**

## ABSTRACT

Human activity recognition is one of the most important tasks in computer vision and has proved useful in different fields such as healthcare, sports training and security. There are a number of approaches that have been explored to solve this task, some of them involving sensor data, and some involving video data. In this paper, we aim to explore two deep learning-based approaches, namely single frame Convolutional Neural Networks (CNNs) and convolutional Long Short-Term Memory to recognise human actions from videos. Using a convolutional neural networks-based method is advantageous as CNNs can extract features automatically and Long Short-Term Memory networks are great when it comes to working on sequence data such as video. The two models were trained and evaluated on a benchmark action recognition dataset, UCF50, and another dataset that was created for the experimentation. Though both models exhibit good accuracies, the single frame CNN model outperforms the Convolutional LSTM model by having an accuracy of 99.8% with the UCF50 dataset.

## I. INTRODUCTION

The advent of technology has brought about a plethora of incredible advancements that can simplify our lives. These advancements have made it possible to recognize human activities through various methods such as object recognition, feature extraction and motion analysis. Human activity recognition has proved useful in different fields, including healthcare, sports training, entertainment, robotics and management. As modern society continues to evolve, the dynamics of family life are changing, with an increasing number of parents facing the challenge of taking care of children and elderly family members at home while they are away for work. This presents significant concerns related to the safety, well-being, and overall care of vulnerable family members who may require assistance or supervision during the parents' absence. Human activity recognition (HAR) has emerged as a promising technology that can help address these challenges by remotely monitoring and providing insights into the activities of children and elderly family members at home.

In the healthcare industry, human activity recognition can be used to monitor the daily activities of patients with chronic conditions like Parkinson's disease. This can help healthcare professionals to identify changes in behaviour or mobility, and adjust treatment plans accordingly. In the field of sports, HAR can be used to analyse athletes' performance and technique, identifying areas for improvement and optimizing training plans. In smart homes, HAR can be used to automate home appliances and devices based on the activities of the occupants, For example, lights can be turned on when someone enters a room. In robotics, HAR can be used to improve the capabilities of robots, allowing them to better understand human behaviour and interact with them effectively. In security, HAR can be used to detect suspicious behaviour or intrusions in sensitive areas, such as airports, banks, or military bases. Overall, HAR has a wide range of applications in various fields and industries, and it is a rapidly growing area of research and development.

There are different approaches that have been investigated before with the aim of solving the problem of human activity recognition. Some of them are based on data collected from sensors [1] while others are based on video data. Most times, approaches using sensor data require the subject in the picture to be wearing some type of wearable sensor on their body. In scenarios such as the home environment, it may not always be possible to wear a sensor, and hence this approach may not be easy to implement in all cases. Another approach is capturing a video with the help of a camera and detecting the action performed based on the video. This approach would be more suitable in scenarios where it would be difficult to wear a sensor. Some papers have explored methods to solve human activity recognition in videos. One such method [2] is based on a hierarchical codebook model of local spatio-temporal video volumes. This method is based on a bag of video words (BOV) representation and does not require prior knowledge about actions, background subtraction or motion estimation. But one major drawback of this method is that such a video representation of activities in a scene cannot be applied for long-term behavior understanding, e.g., behaviors that consist of numbers of activities that occur sequentially. Other approaches such as Support Vector Machines (SVM) [3] and Hidden Markov Models (HMM) [4] have also been tried and tested to see their performance on human activity recognition.

In this paper, we propose two deep learning-based approaches to make predictions about human actions in videos. The first model is a single frame CNN model and the second is a Convolutional LSTM model. We aim to investigate the performance of these models on two datasets. The first dataset that we are using to train and evaluate the model on is a benchmark action recognition dataset, UCF50. The second dataset is a dataset that was created by the authors of this paper for the purpose of this experimentation. This dataset was created with the aim of including more Asian faces in the dataset, as most common human action recognition datasets have lesser representation of Asian faces. The model we build should be able to predict actions in a video where a number of different activities occur sequentially. Furthermore, we aim to compare the results of both the models based on their testing accuracies to determine which model performs better. This paper is organized as follows: in Section II, we review the related literature on HAR ; in Section III, we describe the datasets used; in Section IV, we describe our proposed approach; in Section V, we present and discuss our results and finally, in Section VI, we conclude and provide recommendations for future research.

## II. RELATED WORK

Human activity recognition (HAR) has a vast number of applications in industries such as healthcare, sports, and security. Due to its various applications, several techniques have been explored previously to tackle the task of human activity recognition. Various machine learning algorithms and techniques have been developed to accurately recognize human activities from sensor data, including deep learning, support vector machines, and decision trees. The ultimate goal of HAR is to improve the quality of life by enabling personalized and proactive interventions based on the patterns of human behavior.

Jin Wang, Zhongqi Zhang, Li Bin and Sungyoung Lee (2014) [5] proposed an " An Enhanced Fall Detection System for Elderly Person Monitoring using Consumer Home Networks " where HAR was adopted in surveillance systems installed at public places i.e. banks or airports to prevent crimes and dangerous activities from occurring at public places. The findings confirmed that the proposed approaches can recognize ongoing interactions between two or more humans at the earlier stage.

Abdulhamit Subasi, Kholoud Khateeb, Tayeb Brahimi and Akila Sarirete (2019) [1] proposed a "Human activity recognition using machine learning methods in a smart healthcare environment" using the machine learning techniques. HAR is used widely for monitoring the activities of elderly people staying in rehabilitation centers for chronic disease management and disease prevention. HAR is also integrated into smart homes[19] to track elderly people's daily activities. Besides, HAR encourages physical exercises in rehabilitation centers for children with motor disabilities, post-stroke motor patients, patients with dysfunction and psychomotor slowing, and exergaming.

Djamila Romaissa Beddiar, Brahim Nini, Mohammad Sabokrou and Abdenour Hadid (2020) [6] proposed a "Vision-based human activity recognition: a survey" where human activity recognition has been applied quite commonly in gaming and exergaming, and adults with neurological injury. Through HAR, human body gestures are recognized to instruct the machine to complete dedicated tasks. Elderly people and adults with neurological injury can perform a simple gesture to interact with games and exergames easily. HAR also enables surgeons to have intangible control of the intraoperative image monitor by using standardized free-hand movements. Human activity recognition or HAR is the process of interpreting human motion using computer and machine vision technology. With its full potential yet to be discussed, it has gained a lot of attraction for its usefulness and versatility. Shian-Ru Ke, Hoang Le Uyen Thuc, Young-Jin Lee, Jenq-Neng Hwang, Jang-Hee Yoo, Kyoung-Ho Choi (2013) [7] proposed a "A Review on Video-Based Human Activity Recognition" Any algorithm consists of the four basic steps namely data acquisition for training by external or wearable sensors, pre-processing by noise reduction or windowing, feature extraction as structural and statistical features and classification using machine learning and deep learning models.

Homay Danaei Mehr and Huseyin Polat (2019) [8] proposed a "Human Activity Recognition in Smart Home With Deep Learning Approach" where human activity recognition is broadly classified into vision based and Sensor based which again is into wearable[20], object tagged, dense sensing. Ling Chen, Xiaoze Liu, Liangying Peng and Menghan Wu (2020) [9] in their paper "Deep learning based multimodal complex human activity recognition using wearable devices" presented a survey of existing research work which uses a vision based approach for activity recognition and classified the literature in two main categories: unimodal and multi-modal approaches. Unimodal methods are those which use data from a single modality and are further classified as stochastic, rule based, space-time based, and shape-based methods. Multimodal approaches use data from different sources and are further divided into behavioral, effective, and social-networking methods.

Ong Chin Ann and Bee Theng Lau (2015) [10] proposed a "Human activity recognition: A review" with human activity detection utilizing sensor technologies that has been the subject of extensive research over the last ten years. Accelerometers, motion sensors, biosensors, gyroscopes, pressure sensors, proximity sensors, and other sensors are some of the most popular ones used for activity recognition. Some of the sensors use radio technology, like RFID.

Djamila Romaissa Beddiar, Brahim Nini, Mohammad Sabokrou and Abdenour Hadid (2020) [6] proposed a "Vision-based human activity recognition: a survey" where activity recognition is significantly aided by the application of machine learning. Information can be acquired using a variety of strategies and technologies, but once that information has been gathered, it is the responsibility of an algorithm that uses machine learning to infer or recognize the action. Salvatore Gaglio, Giuseppe Lo Re and Marco Morana (2015) [11] published a "Human Activity Recognition Process Using 3-D Posture Data" where SVM, KNN, Random Forest, Naive Bayes, and HMM are examples of some of the more common machine learning algorithms that are utilized in the process of human activity recognition. Before implementing an algorithm for machine learning, feature selection is another crucial step that must be taken. It's possible that a strong collection of features will produce superior outcomes.

Shugang Zhang, Zhiqiang Wei, Jie Nie, Lei Huang, Shuang Wang and Zhen Li (2020) [12] proposed a "A Review on Human Activity Recognition Using Vision-Based Method" and HAR can also be classified based on techniques as Action based, Interaction based and motion based. Action based is further classified into gesture recognition, posture recognition such as standing, lying, running, cooking, behavior recognition, fall detection as in change of body position from normal to reclining without control, activities of daily living such as eating, sleeping, drinking etc. and ambient assisted living. Adrian Nunez-Marcos, Gorka Azkune and Ignacio Arganda-Carreras(2017) [13] in "Vision-Based Fall Detection with Convolutional Neural Networks" told that motion based activities are significant in terms of presence or absence of a person in itself which is very useful in security and surveillance. RFID leads this area and the solutions are of low cost and high accuracy. This area is classified into tracking, motion detection and people counting.

Liandro B. Marinho, A.H. de souza junior and P.P.Reboucas Filho (2017) [3] proposed a "A New Approach to Human Activity Recognition Using Machine Learning Techniques", the application of machine learning significantly which facilitates the recognition of activities. Once information is gathered using a variety of techniques and technologies, it is the responsibility of an algorithm that employs machine learning to infer or recognise the action. S.U.Park, J.H.Park, M.A.Al-Masni, M.A.Al-Antari, Md.Z.Uddin, T.S.Kim (2016) [14] published a "A Depth Camera-based Human Activity Recognition via Deep Learning Recurrent Neural Network for Health and Social Care Services" and in the process of human activity recognition, some of the more prevalent machine learning algorithms include SVM, KNN, Random Forest, Naive Bayes, and HMM. Before implementing an algorithm for machine learning, it is necessary to select features. It is possible that a strong collection of characteristics will yield superior results. Healthcare for the elderly, intelligent environments, security and surveillance, human-computer interaction, indoor navigation, shopping experiences, etc. are all applications of HAR.

Going through all the research, we come across a significant number of challenges faced. First could be complex activities, which includes composite activity like exercise which consists of multiple simple activities like sitting, standing etc., presence of multiple subjects, concurrent activities as a simultaneous combination of simple activities. Second is the environmental interference and background. With cluttered backgrounds, variation in light intensity, presences of objects in almost human like form are a major challenge to overcome. Third is the security of the whole system. Most of the research is focused on accuracy, cost and scalability that compromises the aspect of security. Fourth is the scarce availability of datasets to begin with for training. Yang Wang and Greg Mori (2009) [15] published "Human Action Recognition by Semi Latent Topic Models" that any problem statement and its proposed solution require a very processed dataset inclined with the range of activities required. Also the datasets available are either of animated type or of foreign faces for which is difficult to process in an Indian environment. With this as our main focus, we are trying to build an authentic dataset including an array of household activities from standing, sitting to sneezing and crying with videos primarily focused on Indian faces and body type and deploy various proposed methods to test and compare the accuracy.

## III. DATASET:

Different types of datasets are available for human activity recognition purposes. Some of these datasets focus on data captured by sensors like accelerometers, gyroscopes and ECG sensors, while other datasets focus on video data. Among the video datasets that are available, some include videos that were staged by actors performing the actions in predetermined surroundings, whereas others include videos that are collected from sources such as YouTube. For our implementation, we have used one benchmark action recognition video dataset known as UCF50 and another dataset that we created.

UCF50 dataset consists of 50 action categories that consist of realistic videos that are taken from YouTube [16]. The dataset is quite challenging due to the significant variations in camera motion, object appearance and pose, viewpoint, object scale, cluttered background, lighting condition etc. Some of the actions that are included are diving, walking with a dog, jump rope and horse riding. The average number of videos per action category is 133.

The dataset that we created consists of 3 basic activities - walking, sitting and jumping. Some videos in our dataset showcase our friends or family performing the actions, whereas the other videos were taken from YouTube and from websites that provide free stock video clips. The dataset consists of primarily Indian faces, due to the nature in which the data was collected. Each action category includes 21 videos, and the duration of the videos range from 3-15 seconds.

## IV. PROPOSED APPROACH:

In this paper, we aim to use deep learning techniques to identify human actions from video data. We implement two models, namely Single Frame Convolutional Neural Networks and ConvLSTM to build the human activity recognition system. We begin with the collection of video data, and then we perform a step to extract frames from the video files and preprocess them. The preprocessed frames are then used to create a dataset with a fixed number of images per class. This dataset is then used for training the model. A simple architecture of the proposed approach is given in Figure 1.

The labels (activities) are One Hot Encoded before proceeding with the training. One hot encoding is a way to represent categorical variables as numerical values, where each unique value is represented by a separate binary feature. This technique is especially useful when dealing with multi-classification problems, like ours. By using one hot encoding, we can ensure that each class is treated independently, and the model doesn't assume any ordinal relationship between the classes. After this, we can proceed with the training of our models. The architecture of the two models that we are using would be explained in detail later in this section. During the training of the model, early stopping callback is employed to monitor the validation loss. While fitting the model, the epochs are set to 50, denoting that the total number of times the training dataset can pass through the network. However, if the validation loss does not improve for a given number of epochs (here we've set that given number to 15), then the early stopping callback will halt the training process early. The best weights are restored when the training stops. Early stopping helps reduce the model training time and prevent overfitting by stopping the epochs before they start overfitting.

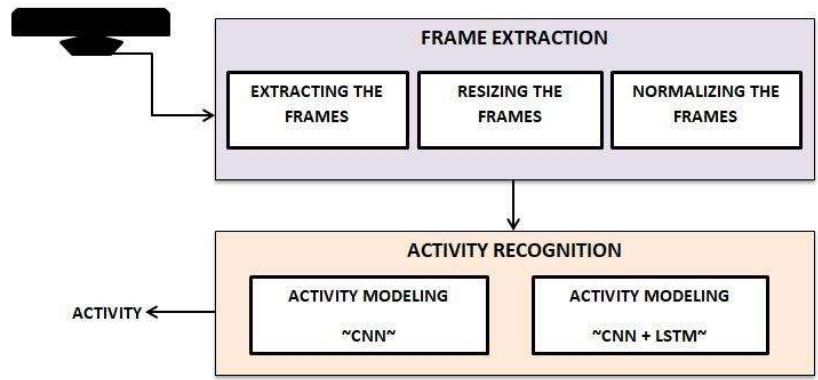

Figure 1. Architecture of the proposed approach

**Data Preprocessing:**

The frame extraction procedure is an essential pre-processing step for video-based applications. Video-based machine learning algorithms require access to the individual frames of a video in order to analyze it. The video frames extraction process involves iterating through all the video files present in the dataset's class directories, reading each video file using the OpenCV VideoCapture method, extracting the video frames and then iterating through each video frame to resize it to fixed dimensions (64x64 pixels) and normalize its pixel values by dividing it by 255[17]. The preprocessed frames are then added to a temporary list for each class. Later we run a function to extract the frames from all the videos from all the action categories. Resizing the frames to a constant image height and image width and normalizing the pixel values to a value between 0 and 1 makes the training easier and is helpful in increasing the performance and accuracy of the model.

**Single Frame CNN:**

CNN, which expands to Convolutional Neural Networks, is a deep learning technique that has been widely used over the years to solve computer vision tasks such as classification of images and object detection. CNNs are inspired from the structure and function of the human visual system and are designed in such a way that they can automatically learn hierarchical representations of input data. CNN has been shown to be more accurate and effective than older methods that relied on manually designed features. At a high level, CNN consists of multiple layers of neurons, pooling layers and fully connected layers. The convolutional layers use learnable filters to scan an input image or video and extract features from it. The pooling layers then reduce the size and retain only the necessary and valuable features that are of significance. The job of the fully connected layers is to then classify the data based on the extracted features. CNN can be used for video classification applications by using the spatial and temporal features extracted from each video frame to make predictions about the entire video sequence.

We are looking for a method that would take a video as input and would give the activity that is being performed in the video as the output. In this single frame CNN approach that we are proposing, we will run a model that would perform image classification on every single frame of the video to recognize the action being performed. The model generates a probability vector for each input video frame, which denotes the probability of different activities being present in that frame. Then we would average all the individual probabilities to get the final output probabilities vector.

Since the videos can be many seconds long, it would not be feasible or necessary to run the classification model on each and every frame. It would be sufficient to run the model on a few frames spread throughout the whole video.

The Convolutional Neural Network (CNN) architecture consists of two convolutional layers, a batch normalization layer, a max-pooling layer, a global average pooling layer, and two fully connected (dense) layers. An empty model object is initialized that can be used to add layers to the model sequentially. The first convolutional layer of the model applies a set of 64 filters (also called kernels) to the input image, where each filter is of size 3x3. The activation function used for this layer is Rectified Linear Unit (ReLU) which is defined as

$$f(x) = \max(0,x)$$

The second convolutional layer of the model also applies a set of 64 filters to the input image, where each filter is of the size 3x3. The activation function used for this layer is also ReLU. A batch normalization layer is added to the model after the second convolutional layer. This layer helps to stabilize the learning process by normalizing the output of the previous layer. A 2D max pooling layer then performs max pooling on the output of the previous layer, with a pool size of (2,2). This reduces the spatial dimensions of the output. A 2D global average pooling layer is then used to take the average of all the values in each feature map, resulting in a single value for each feature map. This reduces the dimensions of the output further. Then a fully connected (dense) layer is added to the model. This has 256 units and uses ReLU activation. Another batch normalization layer is added to the model after the dense layer. Another dense layer is used as the final output layer. This layer has a number of nodes equal to the number of classes in the classification task, which in our case is the number of human actions. The activation function used for this layer is the softmax function, which converts the output of the layer into a probability distribution over the classes.

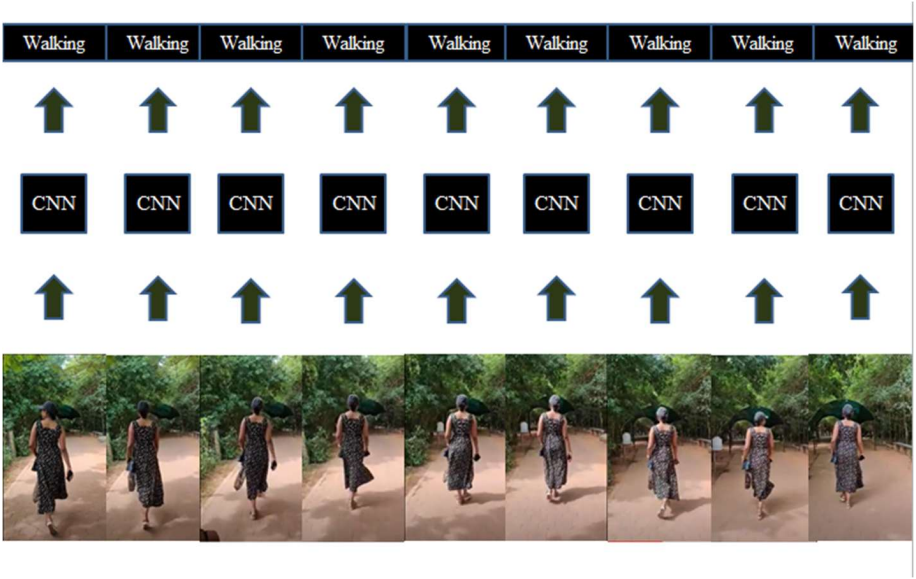

Figure 2. Single Frame CNN

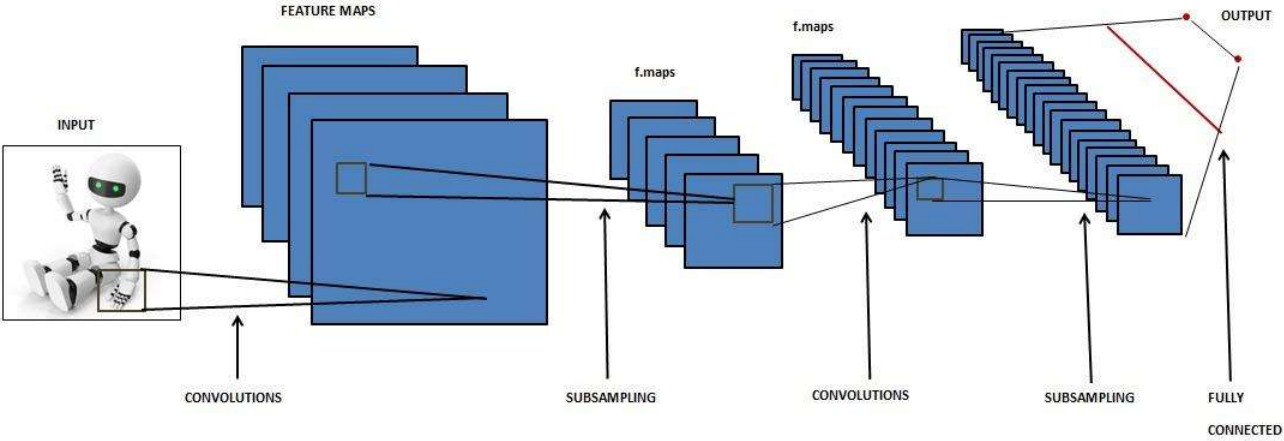

Figure 3. CNN architecture

**Convolutional LSTM:**

Convolutional LSTM (ConvLSTM) is a variant of the Long Short-Term Memory (LSTM) network architecture that is designed to handle sequential data with spatial structures, such as video data[18]. 2D ConvLSTM is a type of Convolutional LSTM that is designed for processing spatiotemporal data in the form of two-dimensional (2D) arrays, such as image sequences or video frames. Traditional CNNs are good at working with image data, hence are suitable for extracting spatial features from individual frames. On the other hand, LSTM networks are good at modeling temporal dependencies, and hence are suitable for working with sequence data. ConvLSTM combines the power of CNN and LSTM networks to effectively capture both spatial and temporal features of the data, hence making it a suitable approach for solving computer vision problems like video classification.

2D ConvLSTM is similar to traditional LSTMs, but with an added spatial convolution operation that is applied on the input and recurrent connections. The input and output of a 2D ConvLSTM cell are both 2D arrays, and the cell can be stacked to form deeper networks that can capture more complex spatiotemporal patterns.

The first layer of the model is a ConvLSTM2D layer that takes input sequences of images with the shape of (sequence length, image image_height, image_width, 3) and applies a 3x3 convolutional filter with 4 output filters and a tanh activation function. The output of the first ConvLSTM2D layer is passed through a 3D max pooling layer with a pool size of (1,2,2), which reduces the spatial dimensions of the output while keeping the number of filters constant. The output of the max pooling layer is passed through a TimeDistributed layer that applies dropout regularization with a rate of 0.2 to each time step of the sequence. The Dropout layer sets a fraction of the input units to zero at each update during training time which helps prevent overfitting. Two additional ConvLSTM2D layers, each followed by a max pooling layer and a TimeDistributed dropout layer, are added to the model with an increasing number of filters. Finally, a Flatten layer is used to convert the output of the previous layer to a 1D tensor. This is followed by a Dense layer with softmax activation, which is the fully connected layer, to predict the class probabilities for each input sequence.

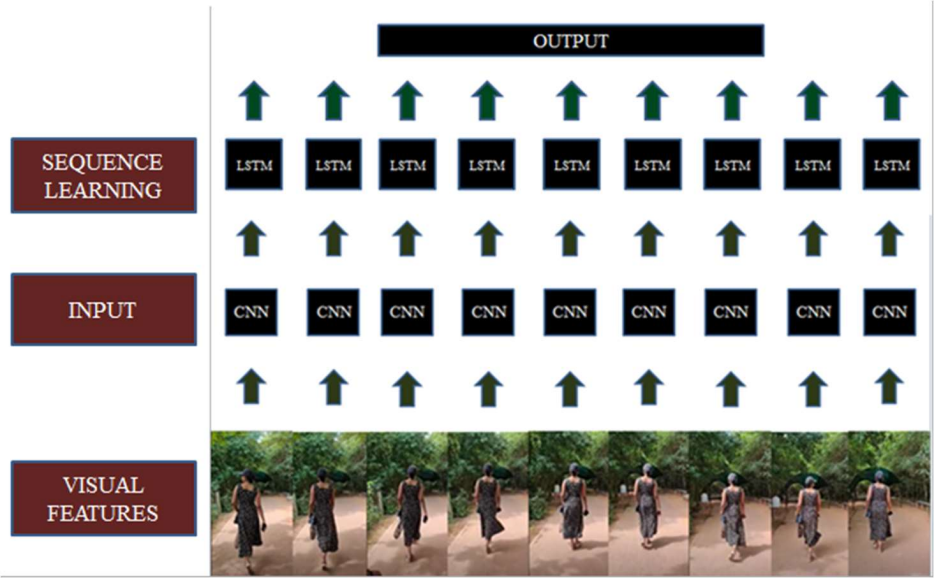

Figure 4. Convolutional LSTM

V. **RESULTS**

During experimentation, we evaluated both the models on the two datasets that we described above: UCF50 and our own dataset. Both the datasets were used individually, and were not combined for experimentation. 80% of the dataset was used for the training stage and 20% of the dataset was used for testing, in both the cases. While fitting the model, the validation split parameter was set to 0.2, hence 20% of the training data was used for validation. The four cases that we have explored are Single Frame CNN with UCF50 dataset, Single Frame CNN with our own dataset, Convolutional LSTM with UCF50 dataset and Convolutional LSTM with our own dataset. For all the four cases, we have plotted the total loss vs. total validation loss graph and total accuracy vs. total validation accuracy graphs, in Fig. . Both these plots can be used to assess the performance of the model during training. The x-axis shows the number of epochs, while the y-axis shows the value of the loss function or the value of the accuracy metric.

For all the four cases, we have recorded the testing accuracy which is given in Table 1. Based on the results, we can see that the single frame CNN model outperforms that Convolutional LSTM model. We can also observe that the accuracy obtained is more with UCF50 dataset than with our own dataset. This is understandable since the dataset we created is small, hence the training size is less. The highest accuracy obtained is for the single frame CNN model with UCF50 dataset, which is 99.8%. We have used a heatmap to represent the confusion matrix for the four cases, for demonstrating the correspondence between the predicted labels, along the x-axis, and the true labels, along the y-axis, and to represent the recognition performance for each action class that was selected. A confusion matrix generally includes 4 groupings: True Positive, which denotes the instances that were correctly identified as positives, False Positive, which denotes the negative examples incorrectly identified as positives, True Negative, which denotes the negative examples that are correctly predicted as negatives and finally False Negative, which denote the positive instances incorrectly predicted as negative. In our case, the diagonal of the confusion matrix or heatmap represents the activities that are correctly recognized. The other cells represent activities that were predicted as some other activity, for example, if a video of 'Jumping' was predicted as 'Walking'. It is to be noted here that, while we used our own dataset for training, we included the whole dataset during the training process since the size of the dataset is small and the dataset includes only 3 activities. However, while training with the UCF50 dataset, we selected 3 activities, namely 'PullUps', 'WalkingWithDog' and 'PlayingGuitar' from the 50 activities in the dataset for training. This was done due to the resource and time constraints, and to reduce the speed of training the model, since the overall size of UCF50 dataset is quite large. In the heatmaps representing either model's performance on UCF50 dataset, 0 denotes the activity 'PullUps', 1 denotes 'WalkingWithDog', and 2 denotes 'PlayingGuitar' on the x-axis and y-axis. In the heatmaps representing either model's performance on our own dataset, 0 denotes the activity 'Jumping', 1 denotes 'Walking', and 2 denotes 'Sitting' on the x-axis and y-axis. By analyzing the heatmaps, we can observe that except for the heatmap of the convolutional LSTM model on our own dataset, all others are showing a really good performance. The heatmap of convolutional LSTM on our own dataset shows that 83.33% of the time, the activity 'Jumping' is mistaken for 'Walking'. The heatmap of single frame CNN on UCF50 dataset shows the best performance, as videos of 'PullUps' and 'PlayingGuitar' were always predicted correctly, and videos of 'WalkingWithDog' were predicted correctly 99.8% of the time. The models were also further tested on videos containing a sequence of actions being performed, and the actions predicted were close to accurate.

|  | Accuracy (%) with UCF50 | Accuracy (%) with own dataset |
| --- | --- | --- |
| Single Frame CNN | 99.80 % | 90.10 % |
| Convolutional LSTM | 98.93 % | 43.75 % |

Table 1. Accuracies obtained with both the models

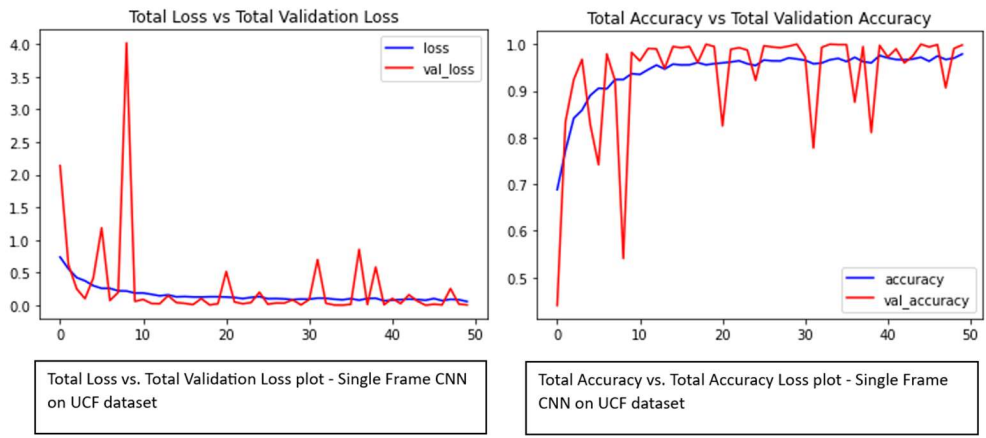

Total Loss vs. Total Validation Loss plot - Single Frame CNN on UCF dataset

Total Accuracy vs. Total Accuracy Loss plot - Single Frame CNN on UCF dataset

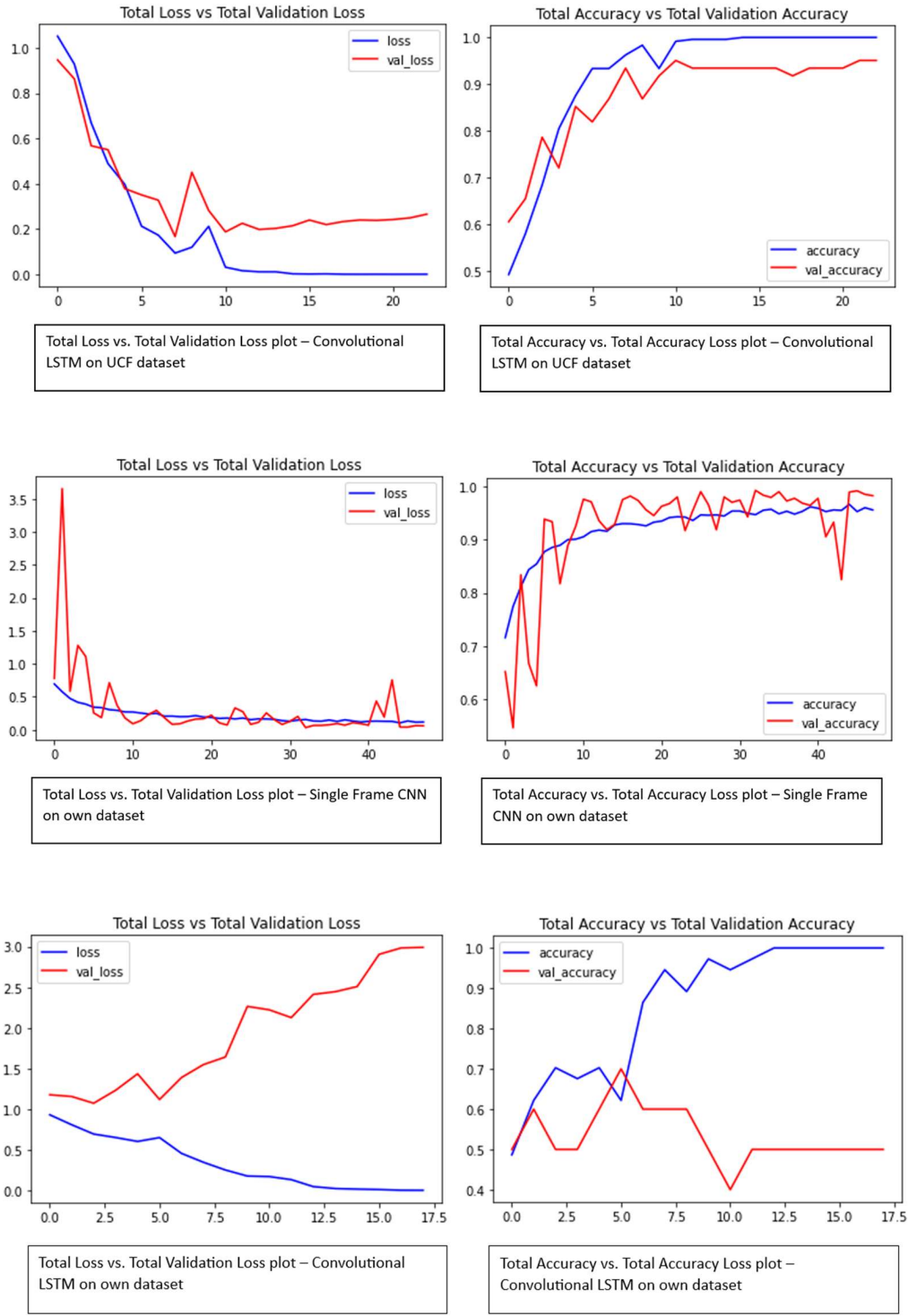

Figure 5. Total Loss vs. Total Validation Loss and Total Accuracy vs. Total Validation Accuracy Plots

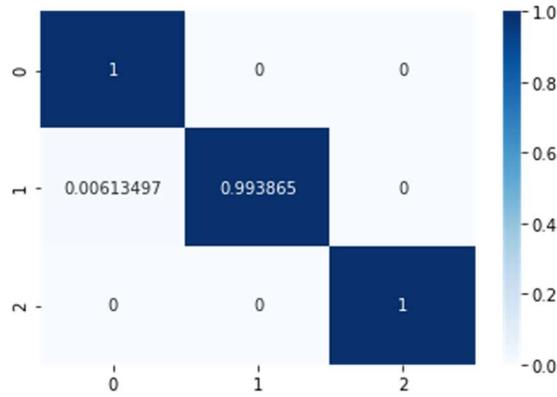
Heatmap of confusion matrix – Single Frame CNN with UCF50 dataset

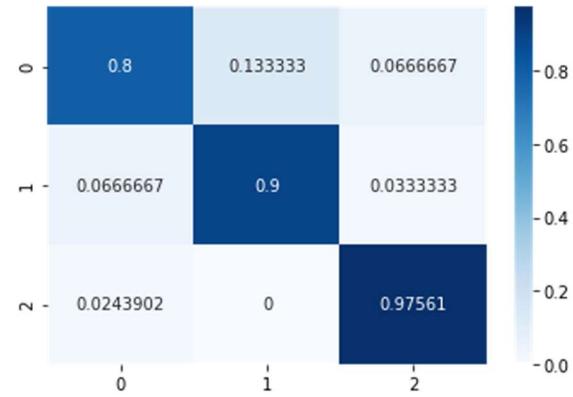
Heatmap of confusion matrix – Convolutional LSTM with UCF50 dataset

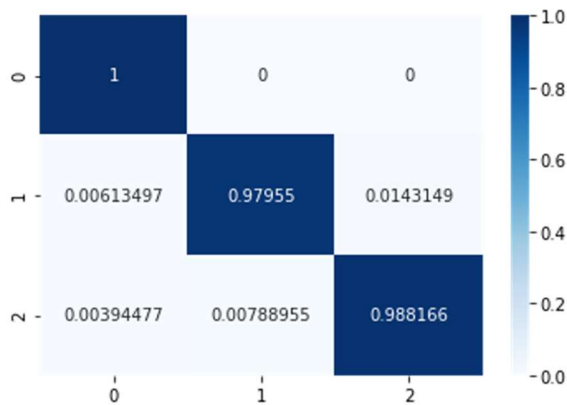
Heatmap of confusion matrix – Single Frame CNN with own dataset

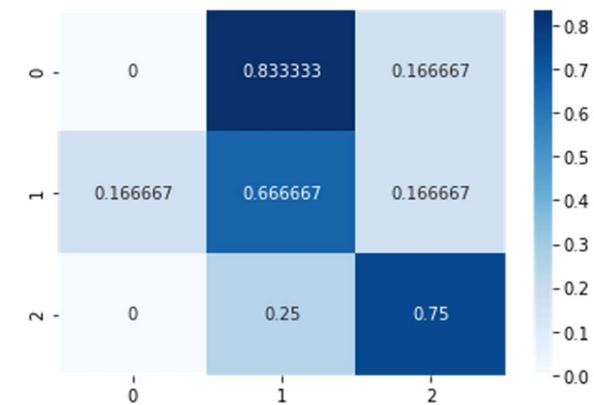
Heatmap of confusion matrix – Convolutional LSTM with own dataset

Figure 6. Heatmaps of confusion matrices

## VI. CONCLUSION

Human activity recognition using single frame CNN and Convolutional LSTM models present a promising solution for prediction of an action in a video the model has not seen before. Traditional CNNs and traditional LSTM networks both have proven to be efficient methods in solving various computer vision tasks. CNN effectively works on image data. LSTM networks on the other hand work well on sequence data. Combining the benefits of both the models makes Convolutional LSTM perfect for video classification, in our case, human activity recognition. Single Frame CNN as well works well for our use case, as CNN can extract features automatically from images. In this study, the two models discussed were trained and tested on UCF50 dataset and a dataset that we created for the purpose of this experimentation. Both the models achieved good recognition performance, however, the single frame CNN model exhibited notably better accuracy than the convolutional LSTM model during testing. Also, the accuracy when testing on the UCF50 dataset was higher for both models when compared to testing on our own dataset. It is understandable as the size of our own dataset is quite small when compared to the size of the UCF50 dataset. For our future works, we would like to augment our newly created dataset to increase the training size and to include more videos of people from different ethnicities. Likewise we would like to expand the proposed models for a larger dataset such as Kinetics 700 because the models can be more effective when applied to a bigger dataset. Further we would like to explore the possibilities of implementing the models in a home monitoring system, by capturing video of activities performed by individuals alone at home in a camera and recognizing the activities done. If dangerous activities such as 'falling' are witnessed, an alert can be sent as an SMS to their caretakers' phones.